\documentclass[runningheads]{llncs}
\usepackage[T1]{fontenc}
\usepackage{graphicx}
\usepackage{booktabs}
\usepackage{multirow}
\usepackage{colortbl}
\usepackage{wrapfig}
\usepackage{amsmath}
\usepackage{amssymb}
\usepackage[stable]{footmisc}
\usepackage{caption}
\usepackage{hyperref}

\begin{document}
\title{VIP: Versatile Image Outpainting Empowered by Multimodal Large Language Model}
\titlerunning{VIP}
%

\author{Jinze Yang \footnotemark[1]\inst{2} \and
Haoran Wang \footnotemark[1]\footnotemark[2]\inst{1} \and
Zining Zhu\inst{2} \and
Chenglong Liu\inst{2} \and
Meng Wu\inst{3} \and
Mingming Sun\inst{4}}
\authorrunning{J. Yang et al.}
%
\institute{Baidu Research, Beijing, China \and
University of Chinese Academy of Sciences, Beijing, China \and
University of Michigan, Ann Arbor, USA \and
Beijing Institute of Mathematical Sciences and Applications, Beijing, China\\
\email{wanghaoran09@baidu.com,yangjinze20@mails.ucas.ac.cn}
}
\maketitle              
\renewcommand{\thefootnote}{\fnsymbol{footnote}} 
\footnotetext[1]{These authors contributed equally to this work.} 
\footnotetext[2]{Corresponding author.} 

\begin{abstract}
In this paper, we focus on resolving the problem of image outpainting, which aims to extrapolate the surrounding parts given the center contents of an image. Although recent works have achieved promising performance, the lack of versatility and customization hinders their practical applications in broader scenarios. Therefore, this work presents a novel image outpainting framework that is capable of customizing the results according to the requirements of users. First of all, we take advantage of a Multimodal Large Language Model (MLLM) that automatically extracts and organizes the corresponding textual descriptions of the masked and unmasked part of a given image. Accordingly, the obtained text prompts are introduced to endow our model with the capacity to customize the outpainting results. In addition, a special Center-Total-Surrounding (C-T-S) decoupled control mechanism is elaborately designed to boost text-driven generation by enhancing the interaction between specific spatial regions of the image and corresponding parts of the text prompts. Note that unlike most existing methods, our approach is very resource-efficient since it is just slightly fine-tuned on the off-the-shelf stable diffusion (SD) model rather than being trained from scratch. Finally, the experimental results on three commonly used datasets, \textit{i.e.} Scenery, Building, and WikiArt, demonstrate our model significantly surpasses the SoTA methods. Moreover, versatile outpainting results are listed to show its customized ability.  Our source code is available at: \textcolor{magenta}{\url{https://github.com/ucasyjz/VIP}}

\end{abstract}
\section{Introduction}

With significant advancements \cite{Yi_2023_ICCV,Pan_2023_ICCV,lu2023painterly,qiu2023diffbfr} in image generation, image outpainting has become a striking topic in recent years. This task extends the content of a given image, requiring generated regions to seamlessly align with original regions. Unlike inpainting \cite{pathak2016context,elharrouss2020image,wang2023imagen,yang2017high}, outpainting is more challenging due to limited contextual information between extrapolated and original image regions. 

\begin{figure}[tb]
  \includegraphics[width=\textwidth]{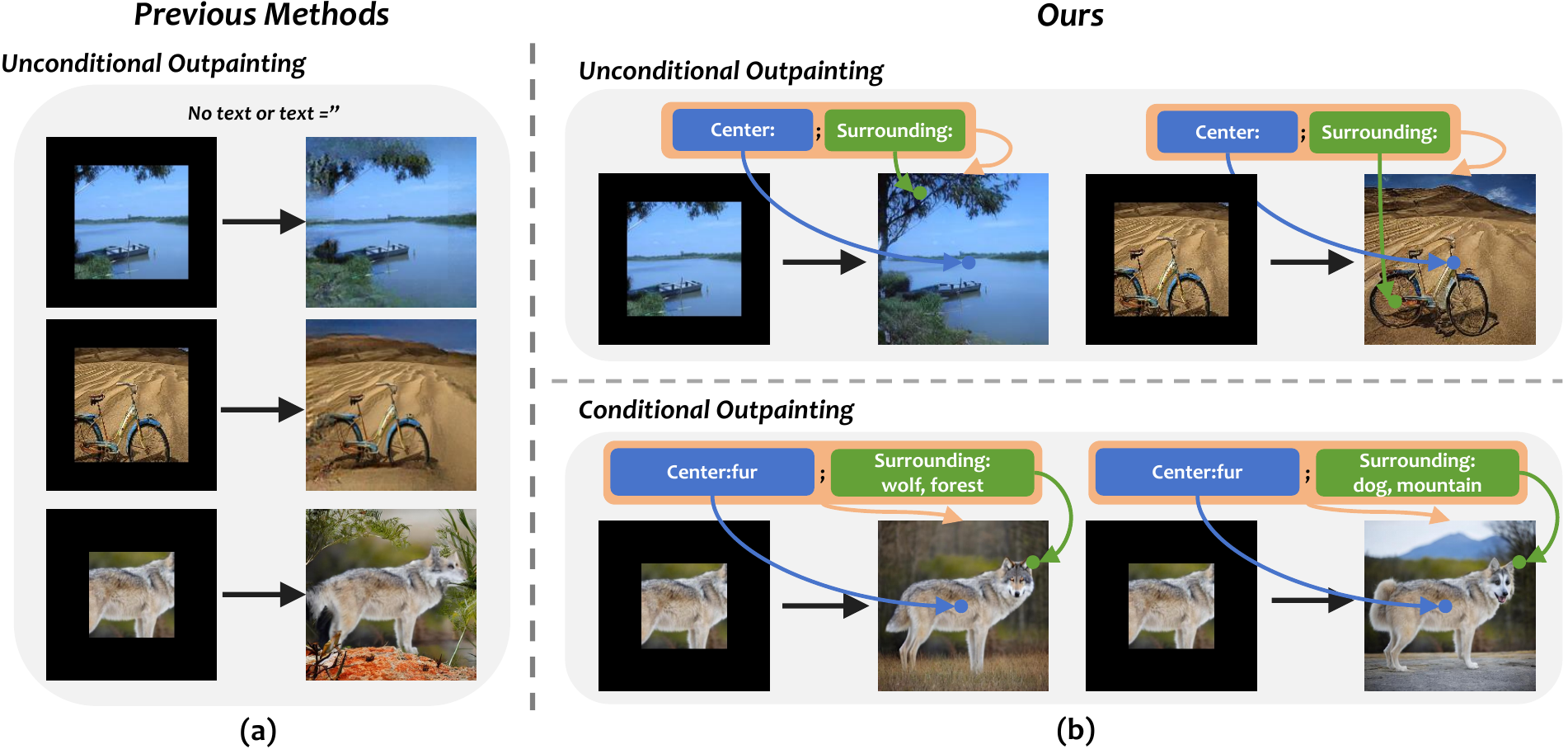}
  \caption{The visualization of versatile outpainting results generated by our proposed method. Unlike existing methods (left), our method (right) can generate satisfactory outpainting results in both unconditional and conditional paradigms.}
  \label{fig:teaser}
\end{figure}

In the early stage, a line of methods is devoted to improving image outpainting via structure modification of the base model. For example, SRN \cite{wang2019wide} introduces semantic content re-generation based on the Convolutional Neural Network (CNN) architecture. Then, U-Transformer \cite{gao2023generalized} is proposed by replacing the CNN structure with the transformer structure for long-range relationship acquisition. To address the issue of object outlines in extrapolated regions not aligning with the input sub-images, Gao et al. \cite{gao2024continuous} exploit eural Ordinary Differential Equations (Neural ODEs) to build a unique bottleneck architecture that enables seamless extrapolation. With its prevalence in numerous visual generation tasks, there have been recent works attempting to introduce diffusion models \cite{ho2020denoising} to resolve the image outpainting problem. PQDiff \cite{zhang2024continuous} learns the correlation between different image regions using the diffusion pipeline.


Although previous works are built based on various architectures and obtain satisfactory results, they typically assume that a given sub-image serves as the sole input for image outpainting, limiting their application and scalability in real scenarios. Generating multiple outpainting results based on various user preferences is more practical than producing a single output automatically. Moreover, many existing approaches are trained from scratch for specific scenes, failing to leverage the full potential of pre-trained large generative models like SD \cite{rombach2022high} and DALLE \cite{pmlr-v139-ramesh21a}. In contrast, fine-tuning these large models for specific tasks has been shown to be effective in various studies \cite{Kawar_2023_CVPR,Karras_2023_ICCV,Everaert_2023_ICCV,Zhang_2023_CVPR,moon2022finetuning}.

To resolve this problem, we present to introduce text prompts as semantic guidance of image outpainting. This approach allows us to customize outpainting results in accordance with our willingness. Moreover, the intervention of text prompts facilitates our exploitation of the well-pre-trained text-to-image generation model \cite{rombach2022high,ramesh2022hierarchical,saharia2022photorealistic}. 
Besides, in existing works \cite{zhang2024continuous,yao2022outpainting} for image outpainting, the relation modeling between different image blocks is conducted in random regions. Actually, we argue there is a nonnegligible distinction between central and surrounding regions of an image. Specifically, in center regions, the content tends to be noteworthy objects or primary information, while in surrounding regions, the content tends to be background information. Considering image outpainting extends from the central regions to the surrounding regions, it is imperative to capture the semantic association between these two spatial positions explicitly.

This paper addresses the above issues by proposing a novel framework namely \textbf{V}ersatile \textbf{I}mage Out\textbf{p}ainting (\textbf{VIP}). First, to enable this model to be aware of the space information through text prompt as query instruction, we need to construct the pairwise prompt-image data. To achieve this, we leverage the Multimodal Large Language model (MLLM) to produce the textual descriptions of the central and surrounding regions of images, followed by organizing the results into a unified format for output. These operations are performed automatically without the need for manual annotation. Based on the obtained image-prompt data pair, to better leverage the capabilities of the pre-trained large model, we construct our image outpainting model modified from an off-the-shelf stable diffusion (SD) model \cite{rombach2022high}. Furthermore, we elaborately design a Center-Total-Surrounding (C-T-S) decoupled control mechanism to capture the fine-grained correspondences between the central and surrounding regions of the image and the corresponding parts in the text prompt, which enhances the customization and generalization capabilities of our model. Notably, unlike existing methods, our model is trained efficient, requiring only fine-tuning on a small amount of data to achieve the current best outpainting effects. As illustrated in Fig. \ref{fig:teaser}, we show that the proposed method not only helps us achieve diverse customization results but also significantly improves the quality of the extrapolated images. 

In brief, the contributions of this paper can be summarized as:
\begin{itemize}
    \item Using a pre-trained prior, we develop a novel diffusion-based image outpainting framework for customized results. Images are annotated with an innovative text prompt format automatically by leveraging MLLM.
    \item A novel Center-Total-Surrounding (C-T-S) decoupled control mechanism is presented to promote the fine-grained interaction between corresponding parts in images and text prompts.
    \item The experimental results not only validate our approach substantially outperforms the existing methods but also can generate diverse results based on customized instructions.
\end{itemize}

\section{Related Work}
\subsection{Image Outpainting} 

The task aims to extrapolate the surrounding contents based on the center contents of the image. The research \cite{sabini2018painting} introduces the deep learning method into the image outpainting task but only performs the horizontal outpainting. IOH \cite{van2019image} is proposed by utilizing the Generative Adversarial Network (GAN) \cite{creswell2018generative} to implement the image outpainting. SRN \cite{wang2019wide} is developed by directly learning the semantic features for outpainting. Kim et al. \cite{kim2021painting} transform the outpainting task into the inpainting task by swapping the outer region of the image with its inner region. Lu et al. \cite{lu2021bridging} propose a bidirectional content transfer module to generate the intermediate region between two different input images. Gao et al. \cite{gao2023generalized} further improve the results with a U-transformer structure and extend the horizontal outpainting to all-side outpainting. QueryOTR \cite{yao2022outpainting} further adopts the ViT module and transforms this task into a patch-wise sequence-to-sequence autoregression problem. Neural ODE \cite{gao2024continuous} is designed by implementing continuous extrapolation in latent space to make the generation results smoother. However, these methods are all trained without text prompt conditions, which suffer from the lack of semantic information guidance. Note that Inout \cite{cheng2022inout} also introduces text as conditions to control the outpainting results. Its text prompt annotation is just represented as one label word. This label is annotated by a semantic segmentation model that only performs close-vocabulary classification on a narrow range of classes. By contrast, our employed MLLM could return the open-world results and automatically organize them into templated answers. Moreover, Inout is built based on GAN architecture rather than diffusion models like ours, which limits its capacity. 

\subsection{Multimodal Large Language Model}

Multimodal Large Language models are designed to process and understand multiple types of data (such as text, images, video, etc.). These models can learn across different data modes and capture associations between different modes to perform well on a range of cross-modal tasks, including Visual Question Answering (VQA), multimodal translation, and more. Here are some important multimodal Multimodal Large Language models. GPT-4 \cite{yang2023dawn} is an open-source MLLM that can understand the image. With the Reinforcement Learning from Human Feedback (RLHF), the feedback from humans enhances the output results. Llama2 \cite{touvron2023llama} is proposed by FaceBook with some improvements in structure and data training. Through three-stage training, Qwen-VL \cite{bai2023qwen} further enhances the visual ability of MLLM by introducing the visual encoder. ERNIE-ViL \cite{yu2021ernie} adopts scene graphs of
visual scenes to build detailed semantic connections across vision and language.

\subsection{Diffusion Model} 
The diffusion model \cite{dhariwal2021diffusion,avrahami2022blended,zhang2024motiondiffuse,bar2024lumiere,Yang_2023_CVPR,Zhou_2023_CVPR} is a generative model for producing high-quality, diverse data, such as images, video, and text. These models generate data by gradually introducing noise into the data and then learning how to reverse the process.

DDPM \cite{ho2020denoising} lays the foundation of the diffusion era for the generation task. It generates data by simulating the diffusion process of data with high quality and diversity. To speed up the diffusion process by reducing computation costs, LDM \cite{rombach2022high} transforms the image space into the latent space through a vae encoder and maps it back to the image space using a vae decoder. Meanwhile, it incorporates the text condition into the pipeline for customized generation. Based on LDM, Stable Diffusion is a commonly-used pre-trained large-scale model, which has extensively promoted the development of the generation field. ControlNet \cite{Zhang_2023_ICCV} is proposed to add conditional control by copying a trainable version based on stable diffusion. DALLE-2 \cite{ramesh2022hierarchical} combines the clip and diffusion model to obtain better text-to-image results. Dreambooth \cite{Ruiz_2023_CVPR} achieves subject-driven generation by fine-tuning the pre-trained stable diffusion model on just a few images of a subject. PQDiff \cite{zhang2024continuous} is proposed by learning the correlation between randomly cropped views for image outpainting based on a diffusion model. However, it still lacks text conditions.

\begin{figure}[t]
  \centering
  \includegraphics[width=\linewidth]{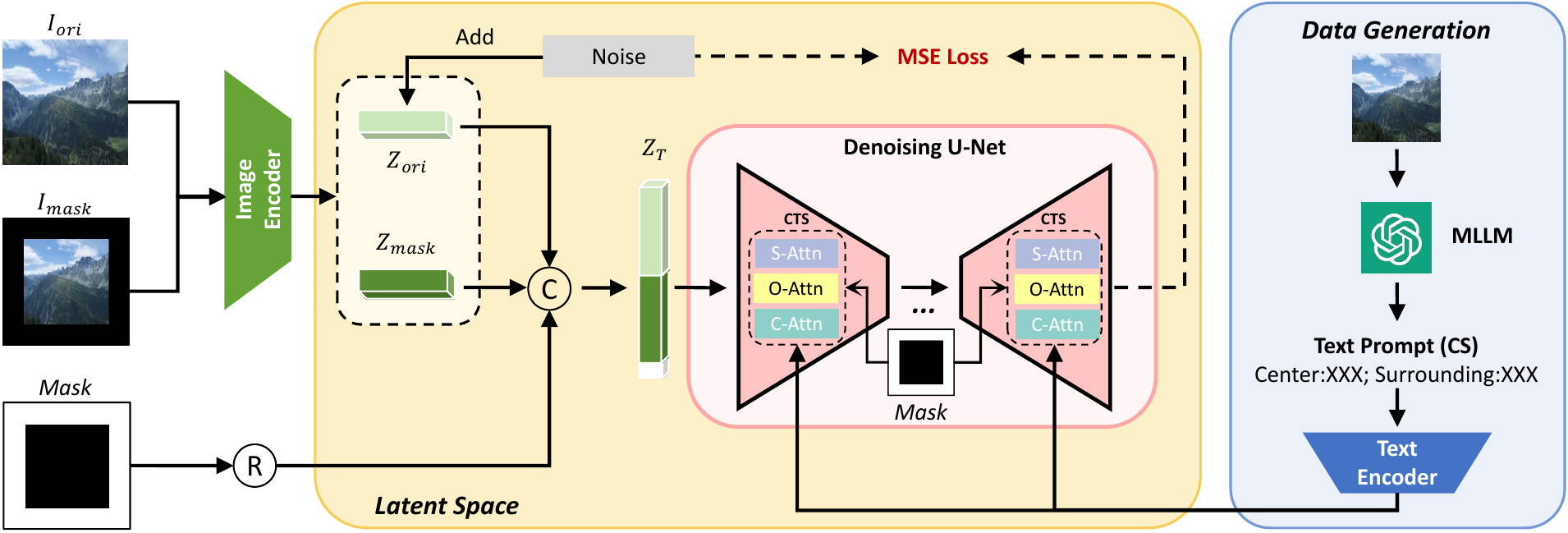}
  \caption{Overall framework of our proposed learning method. First, we concatenate the raw image latent, masked image latent, and mask as the input of the unet. The output will be calculated as mean-square error (MSE) loss with the added noise in the raw image latent. Meanwhile, the text prompt and mask will be input into the proposed C-T-S decoupled control mechanism as conditions. 'R' and 'C' mean the resize and concatenation operations, respectively.}
  \label{fig_whole}
\end{figure}

\section{Methods}
In this paper, we introduce a novel learning framework based on the diffusion model namely versatile image outpainting (VIP), and the whole pipeline is demonstrated in Fig. \ref{fig_whole}. The details of each component are described in the following.

\subsection{Preliminaries}
\textbf{Diffusion Model.} Diffusion Models are a class of deep learning models that generate high-quality data by gradually increasing and then decreasing the noise on the data. This model performs well in the fields of image generation, video generation, text generation, etc. The working principle of the diffusion model can be divided into two main stages: the forward and the reverse process. The forward process is demonstrated as follows:
\begin{equation}
\begin{aligned}
q_t(x_t|x_{t-1}) = N(x_t; \sqrt{1 - \beta_t}x_{t-1}, \beta_tI)
\label{f_FOR}
\end{aligned}
\end{equation}
Then the reverse process to obtain final results is computed by:
\begin{equation}
\begin{aligned}
x_{t-1} = \frac{1}{1-\beta_t}(x_t - \frac{\beta_t}{\sqrt{1-\overline{a}_t}}\epsilon_\theta(x_t,t) )
\label{f_rev}
\end{aligned}
\end{equation}
where $\overline{a}_t = \prod_{s=1}^t(1-\beta_s)$. Finally, the diffusion model is trained using the following loss ${L}_{DM}$:
\begin{equation}
\begin{aligned}
\mathcal{L}_{DM} = \mathbb{E}_{x_0, \epsilon, t}\left[ \|\epsilon - \epsilon_{\theta}(x_t, t)\|^2 \right]
\label{f_difloss}
\end{aligned}
\end{equation}

\textbf{Multimodal Large Language Model.} Multimodal Large Language models (MLLM) are advanced artificial intelligence technique that combines the ability to process different types of data. These models gain the ability to understand and generate content across multiple modals by learning large multimodal datasets. They can understand the relationship between text and images, generate text prompts from images, etc. The users should give an input prompt to the MLLM for usage. For example, 'description of the input image', and the MLLM will become an expert in the image caption field and generate the corresponding description. In this paper, we utilize the GPT-4 \cite{yang2023dawn} model due to its excellent performance.

\subsection{Text Paradigm Incorporated with Space Information}
In this section, the implementation of introducing the text prompt condition into the outpainting task is described. It is considered the traditional prompt format mainly lacks space information, which can not provide accurate condition information for specific image regions. The task needs to restore the surrounding content and keep it consistent with the center content by distinguishing the concrete object contents in different regions. The model should be aware of the precise information about different regions in the generation process. Therefore, incorporating the space information into the text prompt condition is significant. It is well known that the existing Multimodal Large Language models (MLLM) have a strong ability to understand image contents. Therefore, without taking a lot of manual labeling cost, we utilize the Multimodal Large Language model to reconstruct a novel text prompt format, 'Center:xxx; Surrounding:xxx' (CS), for fine-tuning the model. From our perspective, better results on conditional and unconditional image outpainting are expected. We redefine the conditional and unconditional generation based on whether the keywords after 'Center' and 'Surrounding' exist. 

Specifically, we define the task prompt as follows, \textbf{“Please use three English keywords to describe the center and the surrounding of the image, and output in the following format: Center:xxx,xxx,xxx; Surrounding:xxx,xxx,xxx”}. The MLLM takes this prompt as input and auto-regressively captions the different regions of the input image. Then, the data pairs of the enhanced text prompt with space information and the corresponding image are obtained. After generation, we filter the data to reduce the noisy samples and set a part of the data with a prompt 'Center:; Surrounding' without keywords as the unconditional outpainting, and the other parts with keywords as the condition outpainting. Therefore, customized outpainting can be achieved based on the generation of novel text conditions. Then, the corresponding fine-tuning can be performed using the proposed text prompt format.

\subsection{Text-driven Center-Total-Surrounding Decoupled Control}

\begin{figure}[t]
  \centering
  \includegraphics[width=\linewidth]{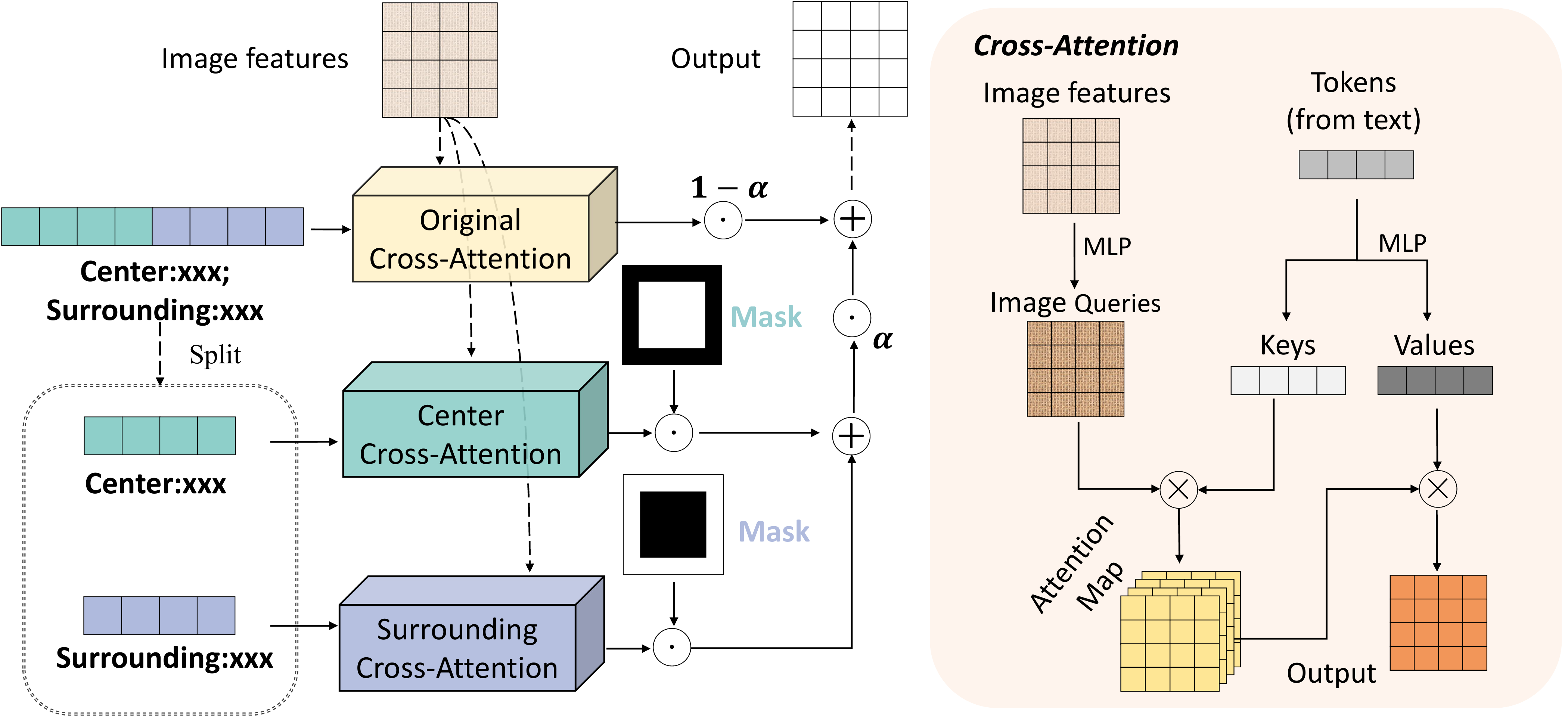}
  \caption{Architecture of Center-Total-Surrounding decoupled control mechanism. '$\cdot$', '$\times$', and '+' mean the element-wise multiplication, tensor multiplication and element-wise addition, respectively. }
  \label{fig1}
\end{figure}

In this section, to make full use of the proposed text paradigm and achieve the alignment between different image regions and the corresponding text parts, the text prompt part for a specific image area should be input into the model, independently. It is expected that different spatial positions of images can be generated by corresponding parts of text control. Therefore, we propose a Center-Total-Surrounding (C-T-S) decoupled control mechanism to realize the purpose as shown in Fig. \ref{fig1}. Specifically, the details of the implementation are described below.

$F_I$ is the latent features of the image, and $F_T$ is the latent features of the text prompt. Under normal circumstances, the cross-attention operation between image and text is calculated as follows:
\begin{equation}
\begin{aligned}
Q & = W_Q \cdot F_I, K = W_K \cdot F_T, V = W_V \cdot F_T  \\
Y &= \text{softmax} \left( \frac{{QK^T}}{{\sqrt{d_k}}} \right) V
\label{f_cross}
\end{aligned}
\end{equation}
Where $W_Q$, $W_K$, and $W_V$ are linear layers to map the two features. 

However, it can be found that the space information is lacking in the original text prompt, and the different text tokens cannot correspond accurately to different image tokens in the masked image without any guidance. Therefore, the C-T-S decoupled control mechanism is proposed to generate the results of the local cross-attention for better conditions. Specifically, we first split the prompt into two parts given a prompt as 'Center:xxx; Surrounding:xxx' according to the center and surrounding settings. Then, we develop two additional cross-attention modules to control the generation of the center and surrounding regions.

We define the text token embedding of 'Center:xxx' and 'Surrounding:xxx' as $F_T^C$ and $F_T^S$, and the mapped features are denoted as $K^C$, $V^C$ and $K^S$, $V^S$, respectively. Then the specific cross-attention between the image and different text parts is computed by:
\begin{equation}
\begin{aligned}
Y^C &= \text{softmax} \left( \frac{{Q{K^C}^T}}{{\sqrt{d_k}}} \right) V^C \\
Y^S &= \text{softmax} \left( \frac{{Q{K^S}^T}}{{\sqrt{d_k}}} \right) V^S
\label{f_new_cross}
\end{aligned}
\end{equation}

The initial weights of the two new attention modules are copied from the original attention module to maintain the strong generation ability of the pre-trained model. Meanwhile, based on the input mask information, we can separate the latent features of the center and surrounding regions. Then, the final results are obtained by adding the two parts:

\begin{equation}
\begin{split}
\hat{Y}^C &= Y^C * (1 - mask)\\
\hat{Y}^S &= Y^S * mask\\
\hat{Y} &= \hat{Y}^C + \hat{Y}^S 
\label{f_newfeature}
\end{split}
\end{equation}

Then we retain the latent features $Y$ obtained by performing cross-attention with the total text prompt to maintain the global consistency. Here, we introduce a trainable parameter $a$ to fuse the two latent features adaptively:
\begin{equation}
\begin{aligned}
Y &= (1-a) * Y + a * \hat{Y}
\label{f_fuse}
\end{aligned}
\end{equation}

Therefore, local conditions for specific regions and global consistency across the whole image can be accomplished. Customized outpainting results are obtained with our proposed method. 

\begin{table}[tb]
  \caption{Comparison with State-of-The-Art methods. The best and second best results are indicated by \textbf{boldface} and \underline{underline}, respectively. '+ copy' means that after the network generates outpainting images, input sub-images are duplicated into corresponding central regions of outpainting images.}
  \label{t_sota}
  \setlength{\tabcolsep}{1.2pt}{
  \begin{tabular}{c|c|cc|cc|cc}
    \toprule
    \multirow{2}{*}{Methods} & \multirow{2}{*}{Type} &\multicolumn{2}{c}{Scenery} \vline &\multicolumn{2}{c}{Building Facades} \vline  &\multicolumn{2}{c}{WikiArt}\\
    \cmidrule{3-8}
     & &FID$\downarrow$  &IS$\uparrow$ &FID$\downarrow$  &IS$\uparrow$ &FID$\downarrow$  &IS$\uparrow$\\
    \midrule
      SRN \cite{wang2019wide} & GAN-based &47.781 
        &2.981 &38.644 &3.862 &76.749 &3.629  \\
      NSIPO \cite{yang2019very} & GAN-based &25.977 &3.059 &30.465 &4.153 
       &22.242 &5.600  \\
     IOH \cite{van2019image}  & GAN-based  &32.107 &2.886 & 49.481& 3.924 & 40.184 
        &4.835 \\
    U-Transformer \cite{gao2023generalized} & GAN-based  &20.575 &3.249&30.542  &4.189&15.904 &6.567 \\
    Neural ODE \cite{gao2024continuous}+copy & GAN-based  &19.352 &3.998  &21.009 & \underline{5.317}  &14.828 &\textbf{8.957} \\
     QueryOTR\cite{yao2022outpainting}+copy & GAN+MAE &20.366 &3.955 &22.378 &4.978 &14.955 &7.896 \\
     PQDiff \cite{zhang2024continuous} & Diffusion-based &29.446 &3.849 &28.855 &4.879 &10.545 &7.374\\
      PQDiff \cite{zhang2024continuous} +copy & Diffusion-based &20.100 &3.981 &\underline{19.133} &\textbf{5.350} &7.968 &\underline{8.605}\\
     \rowcolor[gray]{0.88}
      Ours(GLT) & Diffusion-based  &26.266& \textbf{4.606} &35.054 &4.893 &10.027 & 7.096\\
     \rowcolor[gray]{0.88}
      Ours(GLT)+copy & Diffusion-based  &23.442 &\underline{4.479} &27.158 &5.021 &7.326 &7.437\\
      \rowcolor[gray]{0.88}
      Ours(SFT) & Diffusion-based  &\underline{14.883} & 4.123 & 27.358&5.146 &\underline{3.805} &7.251\\
     \rowcolor[gray]{0.88}
      Ours(SFT)+copy & Diffusion-based &\textbf{10.717} &4.074 &\textbf{17.775} &5.268 &\textbf{3.010} & 7.337\\
    \bottomrule
  \end{tabular}}
\end{table}

\section{Experiment}

\subsection{Dataset}
\textbf{Laion2B-en \cite{schuhmann2021laion,schuhmann2022laion}} The dataset contains about 2B image-text data pairs. We apply the data generation strategy to generate 224 pieces of data with the proposed text prompt format.  \\
\textbf{Scenery \cite{yang2019very}} The Scenery dataset consists of many images of natural landscapes and has about 5,000 images for training and 1,000 images for testing. \\
\textbf{Building Facades \cite{gao2023generalized}} The images in this dataset are focused on urban buildings. There are about 16,000 images in the training set and 1,500 images in the testing set. \\
\textbf{WikiArt \cite{tan2016ceci}} WikiArt is a large dataset of paintings by 195 artists, which contains about 45,500 training images and 19,500 testing images. It has different degrees of realism and stylization.\\
\textbf{MSCOCO \cite{lin2014microsoft}} The MSCOCO dataset features over 330,000 images that depict a wide variety of everyday scenes and objects, including people, animals, and items. Each image is accompanied by five distinct natural language descriptions, providing diverse perspectives on the visual content. In this case, the test set consisting of approximately 40,000 images. 

\subsection{Implementation Details}
In this section, the implementation details are described briefly. Based on the pre-trained stable-diffusion-inpainting model, we perform two training processes. The one, denoted specific training (SFT), is fine-tuning the model on Scenery, Building, and Wikiart dataset for comparison of the SoTA results with the existing methods in the proposed unconditional type, 'Center:; Surrounding:', respectively. As existing methods do, the input size is set to $192\times192$. Unlike the traditional hundreds of epochs for training, the fine-tuning iteration is set to only 30000 with a mini-batch 4. The original images from the datasets are first resized to the set size as ground-truth images. The contents of the region with a size $128\times128$ in the center are preserved, and the surrounding region is masked off. To show the generalization ability of the proposed method and achieve customized results, the other, namely generalized training (GLT), is fine-tuning the model on the data generated by MLLM from the Laion2B-en dataset. With an input size of $256\times256$, the fine-tuning iteration is set to 10000 with a mini-batch 4. A common optimizer, Adam \cite{kingma2014adam}, is selected with a learning rate of 5e-6. In the inference stage, the center size and the extrapolation ratio are set to the same as the corresponding training process. 

\begin{table}[t]
\centering
\caption{Quantitative results under different mask types on MSCOCO}
\label{tabcoco}
\setlength{\tabcolsep}{1mm}{
\begin{tabular}{c|c|c|c|c}
\toprule
\multirow{2}{*}{Method} & \multicolumn{2}{c}{Random Mask} \vline& \multicolumn{2}{c}{Outpainting Mask}\\
\cline{2-5}
 & FID $\downarrow$ & LPIPS $\downarrow$ & FID $\downarrow$ & LPIPS $\downarrow$\\
\midrule
LaMa \cite{suvorov2022resolution} & 27.21 & 0.3137 & 9.52 &\underline{0.1283} \\
SD1.5-inpainting \cite{rombach2022high}&  10.29 &  0.3879  &\underline{5.35} &0.1512  \\
SD3-inpainting \cite{esser2024scaling} & \textbf{8.94} & 0.3465 &$-$ &$-$ \\
SD3-inpainting Turbo \cite{sauer2024fast} & 9.44  & 0.3416 & $-$  & $-$\\
VIP & \underline{9.24} & \textbf{0.2005} & \textbf{1.57} & \textbf{0.1092} \\
\bottomrule
\end{tabular}}
\end{table}

\begin{figure}[t]
  \centering
  \includegraphics[width=\textwidth]{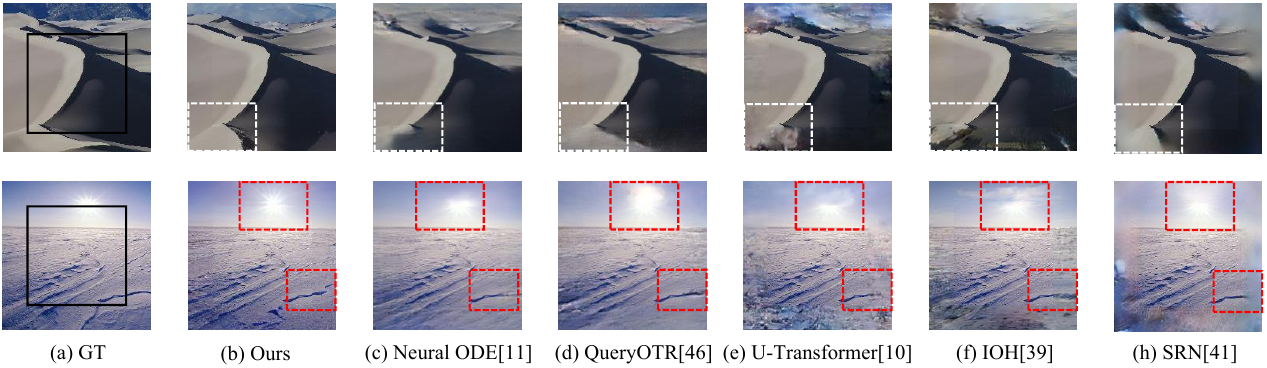}
  \caption{Visualization comparison bewteen different methods on the Scenery dataset. VIP can surpass other SoTA methods in the concerned box region. }
  \label{fig_SOTA}
\end{figure}

For evaluation, the commonly used Frechet Inception distance (FID) 
 \cite{heusel2017gans} and Inception Score (IS) \cite{salimans2016improved} are chosen as the indicators of quantitative performance. 
 
\subsection{Comparisons with State-of-The-Art}
\textbf{Text-Unconditional Generation.} In this section, we compare the outpainting results obtained by the proposed method with those obtained by existing SoTA methods, and the superiority of the proposed method is proven. For quantitative comparison, we select the methods, SRN \cite{wang2019wide}, NSIPO \cite{yang2019very}, IOH \cite{van2019image} , U-Transformer \cite{gao2023generalized}, Neural ODE \cite{gao2024continuous}, QueryOTR\cite{yao2022outpainting} and PQDiff \cite{zhang2024continuous}. The statistical data of FID and IS indicators are recorded in Table \ref{t_sota}. 

As shown in the third and fourth lines from the bottom, performing generalized training (GLT) can achieve a comparable result with the well-behaved methods, indicating that our proposed method has a strong generalization ability and can extend to the outpainting tasks on various image types with more realistic and consistent results. In addition, after performing specific training (SFT) on the three datasets, the results far exceed those of existing methods, as shown in the last two lines. The FID indicator achieves the best level at 10.717, 17.775, and 3.010 on the Scenery, Building, and Wikiart datasets, respectively. Meanwhile, the IS indicator also achieves excellent and advanced results. Especially it can be seen that we do not need to perform fine-tuning on the Wikiart dataset and can accomplish the SoTA FID results just based on the fine-tuning model on the generated data from the Laion-2B dataset.

In addition, the leading results are achieved by just fine-tuning only one-digit epochs instead of fine-tuning the model for a hundred epochs as existing models, which greatly saves training costs based on the strong generation ability of the pre-trained model. Meanwhile, it can be found that whether or not replacing the center content of the generated results with the original center content, our results both exceed the existing best FID result. This means our method can sense the spatial position relation and generate the content of the center and surrounding regions precisely. Moreover, the FID boost effect is not as obvious as the previous method after replacing the center content, which also proves our method can generate whole images with consistent and continuous content. 

We also provide visualization comparison between various methods. As shown in Fig \ref{fig_SOTA}, with our method, the generated content is more clearly visible and consistent in the surrounding regions inside the selected box regions. For example, in the second line, the edge of the dune is smoother and clearer than that of other methods. The outpainting performance has been greatly improved in terms of overall consistency and continuity. 

\begin{figure}[t]
  \centering
  \includegraphics[width=\textwidth]{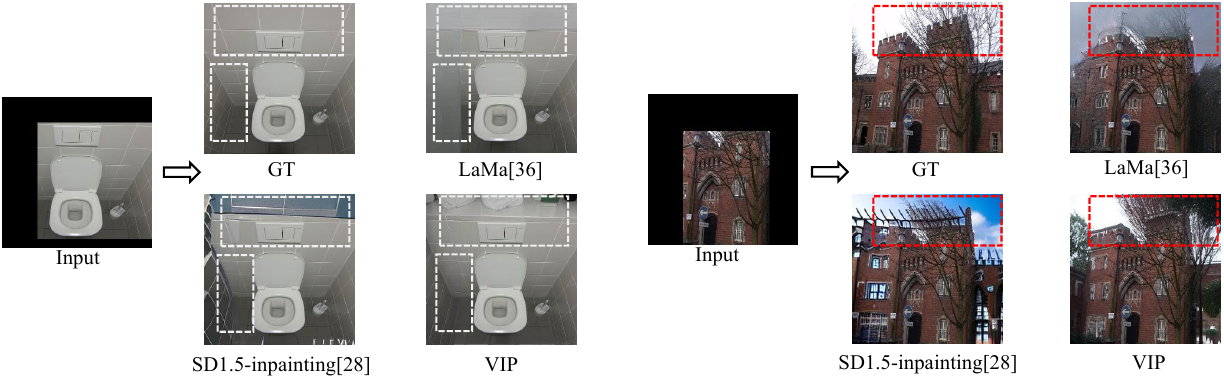}
  \caption{Visualization comparison under different outpainting mask types. }
  \label{fig_diff_outpainting}
\end{figure}

\textbf{Text-Conditional Generation.} To further demonstrate the superiority of the proposed method in text-conditional generation, as done in SD3-Turbo \cite{sauer2024fast}, the quantitative comparison on MSCOCO under random mask types is provided in Table \ref{tabcoco}. In addition, we also provide results only under the outpainting mask type. It can be seen that the performance of VIP far exceeds that of SD1.5-inpainting, especially in outpainting, and reaches a level comparable to SD3 which uses a complex model structure. In addition, the visualization comparison is also demonstrated in Fig. \ref{fig_diff_outpainting}. For different outpainting positions, the performance of our method can exceed that of LaMa and SD1.5 inpainting, in terms of the continuity and reasonableness.

\subsection{Ablation Study}
In this section, we analyze the effectiveness of each component in VIP through various ablation experiments on the test set of the Scenery dataset \cite{yang2019very}. We set the outpainting ratio to 2.25x, like most works. The size of the input center region is 128x128, and the size of the output extrapolation image is 192x192. Both qualitative and quantitative results are provided below.

\begin{table}[t]
\centering
\begin{minipage}[b]{0.47\linewidth}
\centering
 \caption{Effectiveness of each component}
  \label{t1}
  \setlength{\tabcolsep}{3mm}{
  \begin{tabular}{c|cc}
    \toprule
    Method &FID $\downarrow$ & IS $\uparrow$\\
    \midrule
    Baseline   &178.58  &\textbf{4.819} \\
    w/ CS       &16.827   &4.031\\
    \rowcolor[gray]{0.88}
    w/ CS + C-T-S & \textbf{14.883} &\underline{4.123}\\
    \bottomrule
  \end{tabular}}
  
\end{minipage}
\hfill
\begin{minipage}[b]{0.47\linewidth}
\centering
  \caption{Comparison of using different unconditional paradigms}
  \label{t_dup}
  \setlength{\tabcolsep}{3mm}{
  \begin{tabular}{c|cc}
    \toprule
    Method &FID $\downarrow$ & IS $\uparrow$\\
    \midrule
    prompt='\phantom{}' & \underline{15.421} &\textbf{4.131}\\
    \rowcolor[gray]{0.88}
    Ours & \textbf{14.883} &\underline{4.123}\\
    \bottomrule
  \end{tabular}}
\end{minipage}
\end{table}

\begin{figure}[t]
  \centering
  \begin{minipage}[b]{0.47\textwidth}
    \centering
    \includegraphics[width=\textwidth]{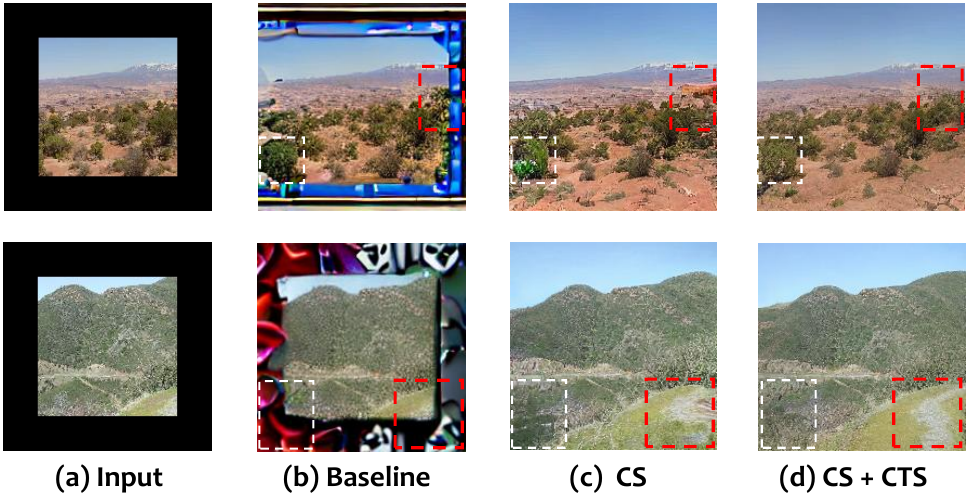}
    \caption{Visualization comparison of using each component.}
    \label{fig_AR}
  \end{minipage}
  \hfill 
  \begin{minipage}[b]{0.47\textwidth}
    \centering
    \includegraphics[width=\textwidth]{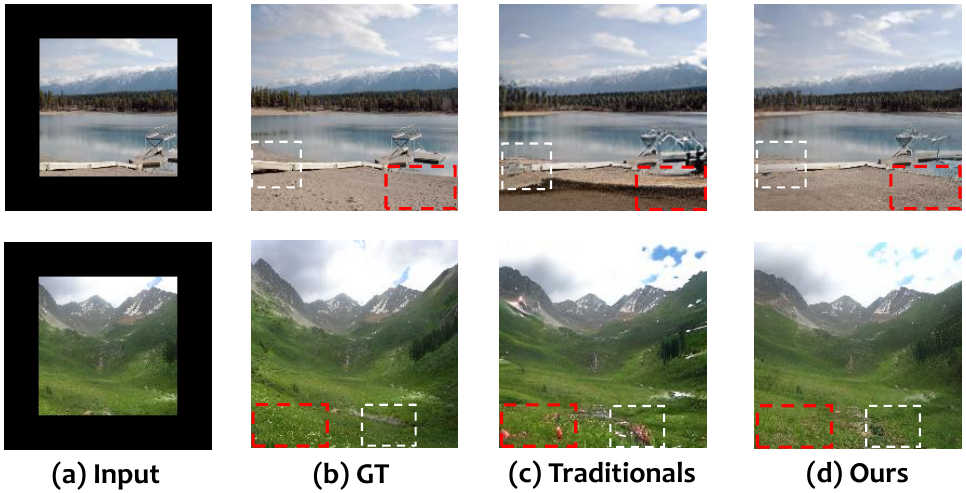}
    \caption{Visualization comparison with different unconditional paradigms.}
    \label{fig_CR}
  \end{minipage}
\end{figure}

\textbf{Effectiveness of Each Component.} To show the contribution of each component in our method, quantitative results are demonstrated in Table \ref{t1}. It is evident that the contribution of each component to the final success of our proposed method is important. The first line shows results obtained by directly using the pre-trained stable-diffusion-inpainting model. The unsatisfactory results demonstrate that the outpainting task is fundamentally different from the inpainting task, and the model optimized for inpainting can not be directly used in the outpainting task for satisfactory results. After fine-tuning the model on the proposed image-text data pair with space information, 'Center:; Surrounding' (CS), the FID indicator has been greatly improved, as shown in the second line. The IS indicator can also maintain a comparable level. Meanwhile, the effectiveness of C-T-S decoupled control mechanism is also verified by further using it, and the results are shown in the third line. The FID and IS indicators are both improved compared to those obtained by only using the proposed text prompt format. This means the introduction of space information helps the model perceive the unique properties of original and extrapolated regions.

We further analyze the effectiveness of each component through the qualitative results as shown in Fig. \ref{fig_AR}. As we can see, the baseline can not behave well in the outpainting task without fine-tuning. After fine-tuning with the proposed prompt format, the outpainting results are obviously improved. Moreover, adding the C-T-S decoupled control mechanism further improves the texture details and reduces abnormal generations. 

\textbf{vs. Traditional Unconditional Prompt Format.}  In this part, we compare the fine-tuning results obtained by the traditional unconditional prompt format, \textbf{prompt = '\phantom{}'}, and the proposed unconditional prompt format, \textbf{'Center:; Surrounding:'}. The results are recorded in Table \ref{t_dup}. It can be found that our proposed format can achieve better results in the FID indicator with a negligible decrease in the IS indicator. The visualization of the results is also demonstrated in Fig. \ref{fig_CR}. It can be found that our method is better than the traditional unconditional format regarding reality and consistency. The performance of the unconditional outpainting results has been improved after utilizing the proposed method.
\begin{wraptable}{r}{0.4\textwidth} 
\centering
        \caption{Results of using different captions for training}
        \label{tab:freq}
        \setlength{\tabcolsep}{1pt}{
        \begin{tabular}{c|c}
            \toprule
            Method & Clip Score \\
            \midrule
            w/ general caption & 28.2435\\
            w/ CS & \underline{29.4943}\\
            w/ CS + C-T-S & \textbf{30.3588} \\
           \bottomrule
        \end{tabular}} 
\end{wraptable}

\textbf{Analysis of Text-Image Consistency.} In this section, we demonstrate the superiority of VIP in terms of the precise text condition by evaluating the clip score as shown in Table \ref{tab:freq}. As shown in the second line, just using CS, the proposed text caption with space information, can improve the text-image consistency. In addition, after using the C-T-S decoupled control mechanism, a better clip score is obtained by enhancing control of text over corresponding image regions.

\subsection{Various Customized Results}

Customized outpainting can be more easily achieved based on text prompt conditions after using the proposed method. In this section, the fine-tuned model by performing the generalized training (GLT) is utilized to achieve customized outpainting. Then we select two samples randomly for visualization. Based on the proposed method, we can achieve customized outpainting results according to the proposed text prompt condition.

\begin{figure}[t]
  \centering
  \includegraphics[width=\linewidth]{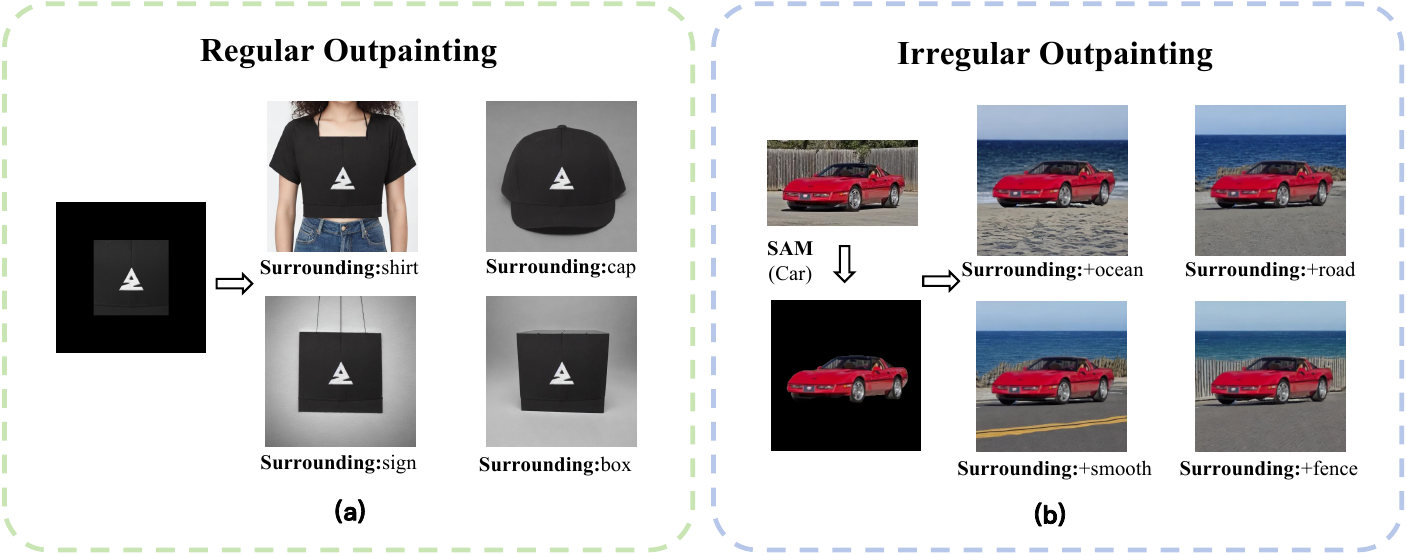}
  \caption{Visualization comparison of customized outpainting results with different text prompts in diverse scenes.}
  \label{fig_7}
\end{figure}

For regular outpainting as shown in Fig \ref{fig_7} (a), it can be found that generated results change after adding different surrounding keywords. If we set the keyword of the surrounding region to 'shirt', 'cap', 'sign', and 'box', generated results correspondingly change to the accurate object. Therefore, precise conditional customized outpainting is achieved through our proposed method. In addition, we can also utilize the MLLM to imagine the surrounding content and output more abundant keywords instead of an artificial mode.

In addition, as demonstrated in Fig. \ref{fig_7} (b), we also perform the customized outpainting with an irregularly shaped mask to show the generalization ability. The noteworthy object in the center region often has an irregular shape. Therefore, we utilize the Grounded SAM \cite{ren2024grounded} to extract foreground objects and mask the surrounding region. Then we obtain the masked image and the corresponding mask to perform the irregular outpainting task. According to elements in the provided prompt, we continuously add the surrounding content with 'ocean', 'road', 'smooth', and 'fence', and the corresponding content will appear in the generated result. Moreover, as we continue to add keywords, the generated result maintains good consistency and continuity based on the original result. Therefore, our model can not extrapolate the region of common square shape like the existing methods but also extrapolate the region of irregular shape. Our model can adapt to various outpainting scenarios and obtain satisfactory results.

\section{Conclusion}
In this paper, we propose a novel framework namely \textbf{VIP} to complete the image outpainting task with the Multimodal Large Language model. The method can introduce text prompts with the space information of the center and surrounding regions to perform precise outpainting. With the help of the Multimodal Large Language model, we design a new text prompt format called 'Center:; Surrounding:' (CS) to assign different object components to specific regions. Then, the proposed Center-Total-Surrounding (C-T-S) decoupled control mechanism can further enhance the correlation between each image region and the corresponding text part through an adaptive fusion of multiple cross-attention operations. Therefore, the model can be aware of the exclusive characters of different regions and generate consistent results across the whole image. We carry out extensive experiments on widely-used datasets. Leading quantitative and qualitative results are obtained in both unconditional and conditional outpainting tasks. The effectiveness of our method is verified. In future work, we intend to continue to explore a more effective text prompt format for customized outpainting results.

\bibliographystyle{splncs04}
\bibliography{main}




\newpage

\titlerunning{VIP}
%

%

In this supplementary material, we provide additional details omitted in the main manuscript due to the limited space. As for implementation details, we first describe the experiment environment and settings and the usage of the Center-Total-Surrounding (C-T-S) decoupled control mechanism. Then, more experiments are conducted to show the effectiveness of the proposed approach, containing a comparison with an open-source outpainting method and an analysis of more different hyperparameter settings. Finally, the visualization comparison with other SoTA methods on more examples is demonstrated. Moreover, various outpainting results obtained by our proposed method are provided below.

\section{More Implementation Details}

\subsection{Training Details}
In this paper, we utilize the pretrained stable diffusion inpainting with an input channel of the unet as 9, which has been fine-tuned explicitly for inpainting. In the fine-tuning process, the whole parameters of the unet are optimized, and other parts like the text-encoder and the vae are frozen. Our sampler utilizes the DDIM sampler with 50 sampling timesteps. In the training stage, there is only a traditional loss function for calculating the mean-square error (MSE) between the predicted noise and added noise. The experiments are carried out on the NVIDIA V100 with Pytorch 1.13.

\subsection{Center-Total-Surrounding Decoupled Control}
For the Center-Total-Surrounding (C-T-S) decoupled control mechanism, we introduce it in the downsampling, middle, and upsampling processes, where the original cross-attention module exists. After calculating the resnet layer, the three cross-attention modules are calculated and the corresponding results are adaptively fused for the subsequent forward process.

\subsection{The Examples of The Generated Training Data}
\begin{figure}[t]
  \centering
  \includegraphics[width=\textwidth]{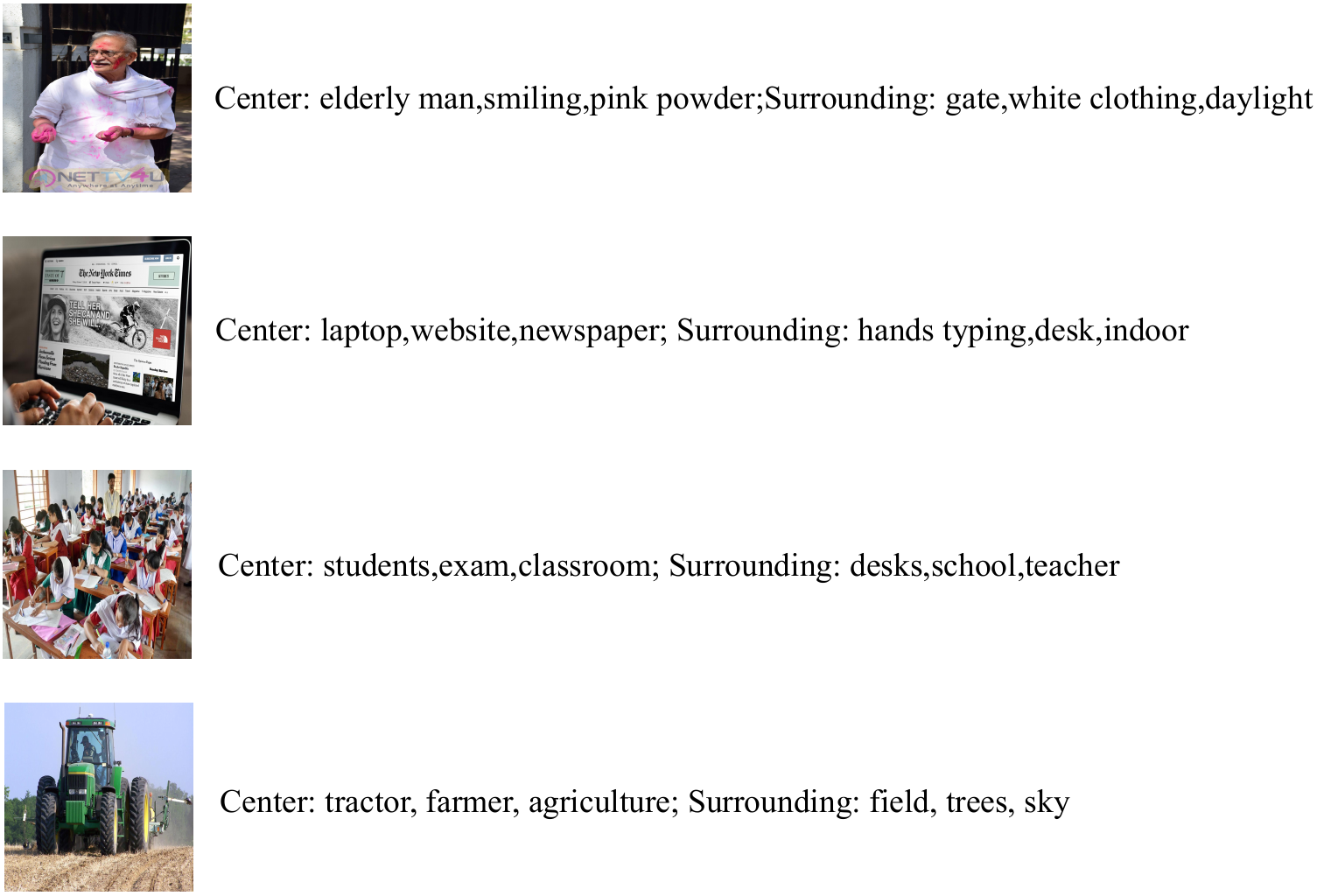}
  \caption{Visualization of samples generated by MLLM.}
  \label{fig_sample}
\end{figure}

In this section, we will demonstrate some examples of the training data generated by MLLM. The selected image-text pair samples are visualized in Fig. \ref{fig_sample}. In the future, we will release the whole generated data. 

It can be seen that the content keywords of the center and surrounding regions are effectively split in the four generated samples. With our proposed approach, we can fully use data in this format for better outpainting performance.

\section{More Experiments}
\begin{figure}
	\centering
	\includegraphics[width=\textwidth]{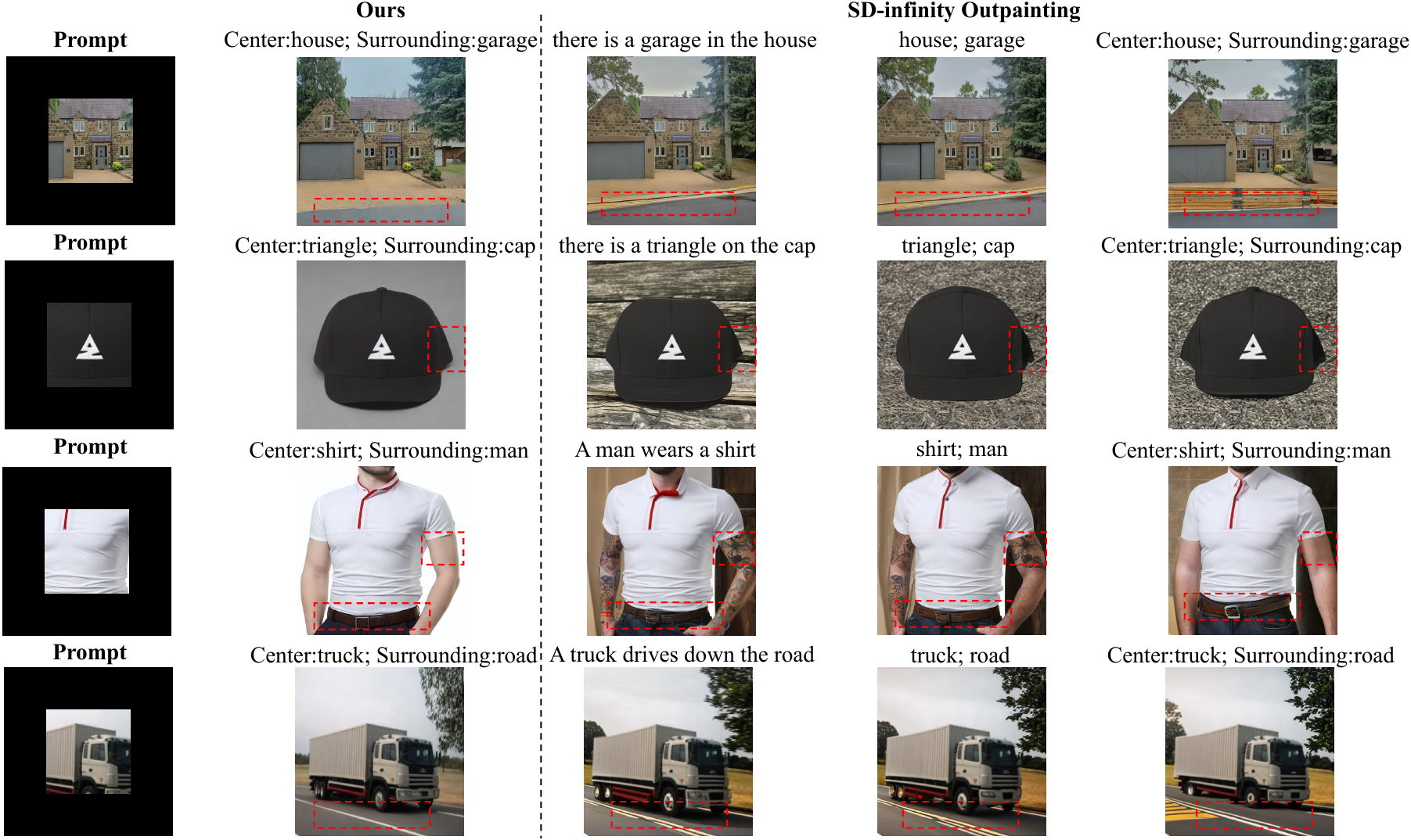}
	\caption{Customized results compared with the open-source outpainting method, SD-infinity Outpainting. \href{https://github.com/lkwq007/stablediffusion-infinity}{}}
	\label{fig_ccomp}
\end{figure}

\subsection{Comparison of Customized Results}
In the main manuscript, we have made a comparison with the leading methods reported in the published paper. To further show the advantage of our work, we further compare it with an unpublished but open-sourced method \footnote{https://github.com/lkwq007/stablediffusion-infinity}, namely \textbf{SD-infinity Outpainting} here, which is also based on the pretrained stable diffusion of the traditional prompt. With the same setting of the sampling parameters, the visualization results are demonstrated in Fig \ref{fig_ccomp}. 

It can be found that after using the proposed approach, the best outpainting results are obtained, and the proposed text prompt paradigm with just some keywords is relatively simple compared to the traditional prompt paradigm with a complete sentence. In the first line, the dividing line between sand and road is clearer and more rational. In the second line, the cap generated by our approach has a more normal shape and a more continuous color. In the third line, the waist area is more realistic, and there are fewer tattoos on the arms In the last line, the generated lane obtained by the open-source outpainting repository is more disorganized.  

\begin{table}[tb]
  \caption{Results of using different parameter formats}
  \label{t1}
  \centering
  \setlength{\tabcolsep}{8.5mm}{
  \begin{tabular}{c|cc}
    \toprule
    Method &FID $\downarrow$ & IS $\uparrow$\\
    \midrule
    baseline & 15.421 &4.131\\
    RIF   &15.170  &\underline{4.216} \\
    CF ($a = 0.5$)  & 15,270  &\textbf{4.224}\\
    \rowcolor[gray]{0.88}
    LF & \textbf{14.883} &4.123\\
    \bottomrule
  \end{tabular}}
\end{table}

\subsection{Analysis of Different Hyperparameter Setting}
In this section, we analyze the influence of setting the hyperparameter $a$ used in the C-T-S decoupled control mechanism. Specifically, we propose three variations for comparison: the random initialization format (RIF), the constant format (CF), and the used learnable format (LF). The corresponding results on scenery are recorded in Table \ref{t1}.

\begin{figure}[t]
  \centering
  \includegraphics[width=\textwidth]{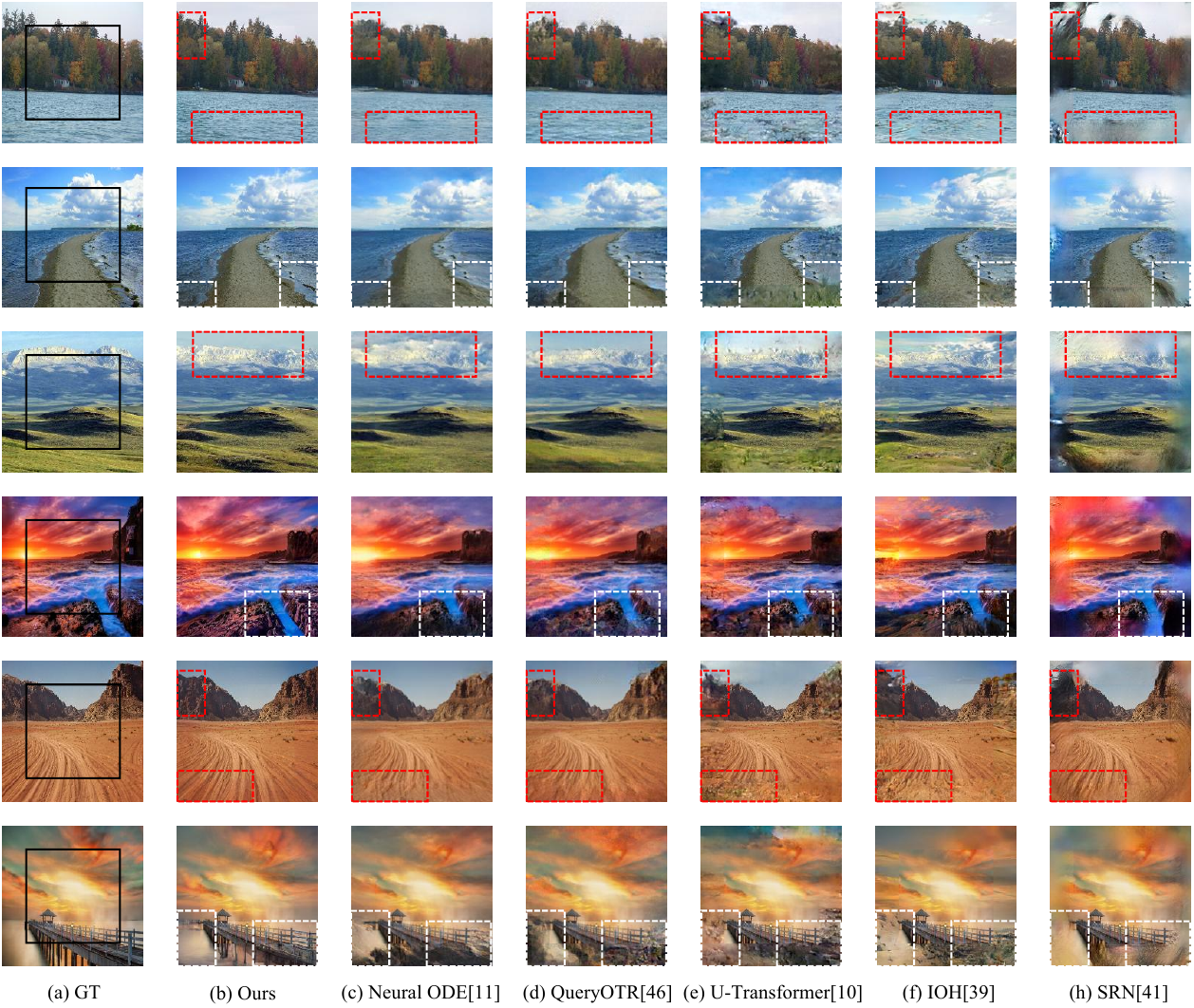}
  \caption{Visualization of more samples compared with other methods on the Scenery dataset.}
  \label{fig_SOTA}
\end{figure}

It can be found that no matter which format we use, the FID results obtained are always better than those obtained by the baseline, and the IS indicator maintains an excellent level. This indicates that our method is insensitive to the parameter format. The best FID results are achieved by utilizing the learnable format compared to the other two formats. It is understandable that directly setting a constant or random initialization will not adapt to the fine-tuning process and obtain better results. Moreover, using the random initialization format is better than the constant format, which means that using the constant format needs to perform extensive experiments to explore the best value, which is very consuming. Therefore, the learnable format is a reasonable and effective strategy.

\section{More Results}
\subsection{Traditional Unconditional Results}

In this section, we provide the visualization of more cases for comparison with other well-behaved methods. The unconditional outpainting results on the Scenery dataset are demonstrated in Fig. \ref{fig_SOTA}. For input, only the part in the black box area of the GT image is reserved, and the other parts will be extrapolated. As we can see, the texture details are more pronounced around the edges, and the contents are more continuous and not as abnormal as in other methods. For example, the reefs and currents in the fourth line have more structural information.

\begin{figure}
  \centering
  \includegraphics[width=\textwidth]{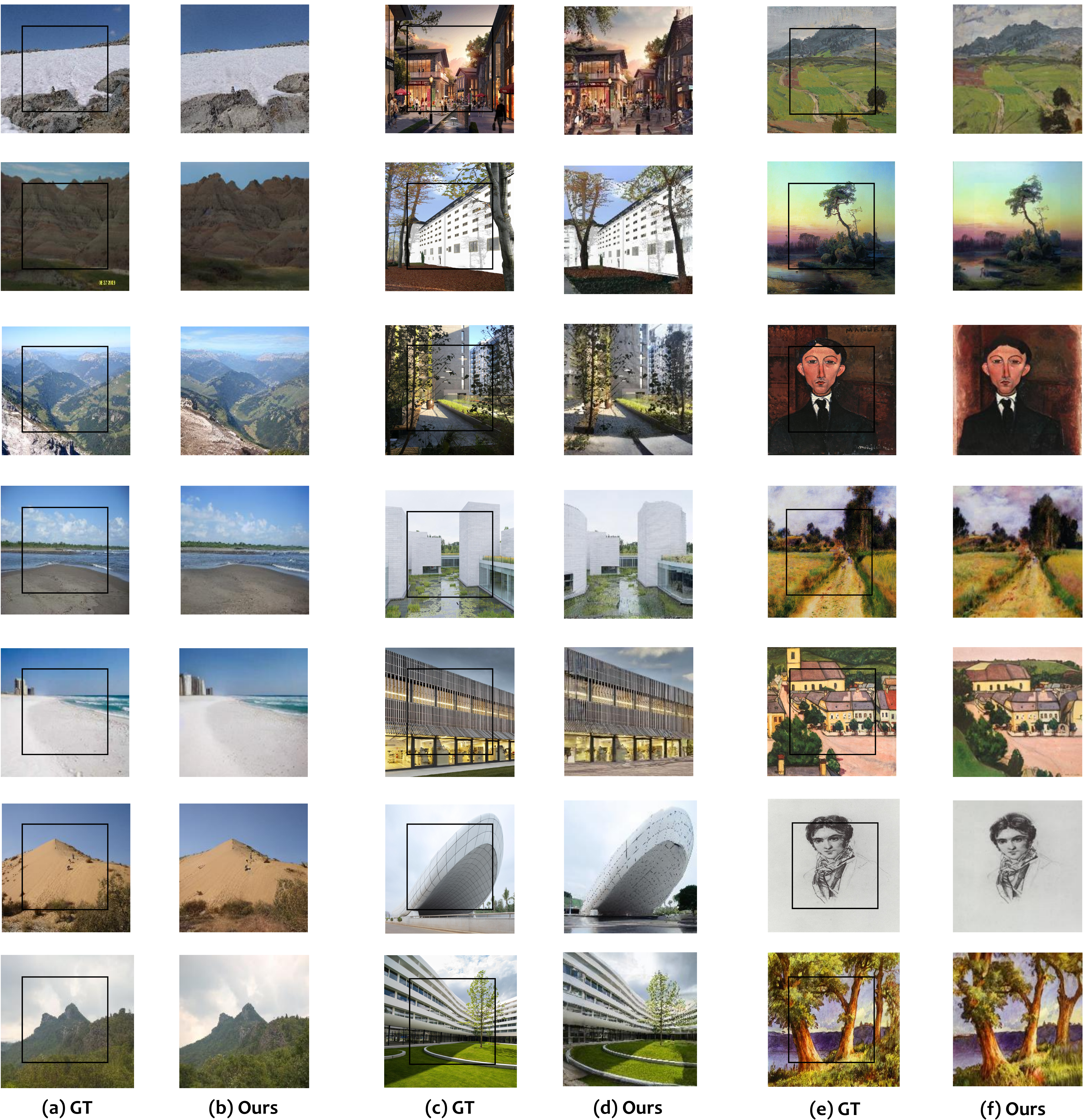}
  \caption{Visualization of more outpainting results on the Scene, Building Facades, and WikiArt datasets.}
  \label{fig_un}
\end{figure}

In addition, we provide more outpainting results generated by our method on the three datasets in Fig. \ref{fig_un}. Left, middle, and right are the Scenery, Building Facades, and WikiArt datasets, respectively. For input, only the part in the black box area of the GT image is reserved, and the other parts are extrapolated. It can be found that the extrapolated part is consistent with the input part, and the outpainting performance is satisfactory on various samples.

\begin{figure}[t]
  \centering
  \includegraphics[width=\textwidth]{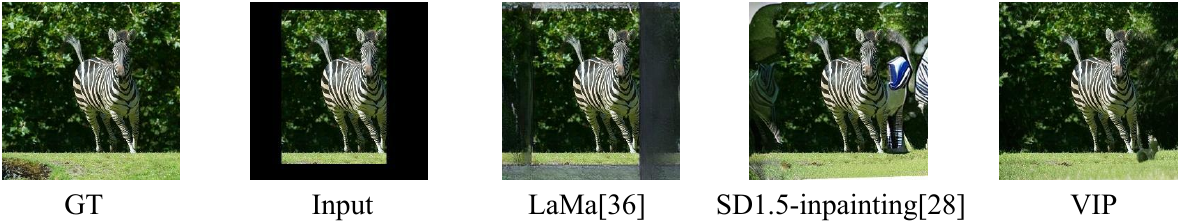}
  \caption{Visualization comparison under non-centered outpainting mask types.}
  \label{fig_diff_100}
\end{figure}

Meanwhile, under masks of different shapes and positions, results on more image types, like cartoon/composite images, are demonstrated in Fig. \ref{fig_diff_100} and the right part of Fig. \ref{fig_other}. It can be found that the generalization of our model is satisfactory.

\subsection{Customized Conditional Results}
\begin{figure}[t]
  \centering
  \includegraphics[width=\textwidth]{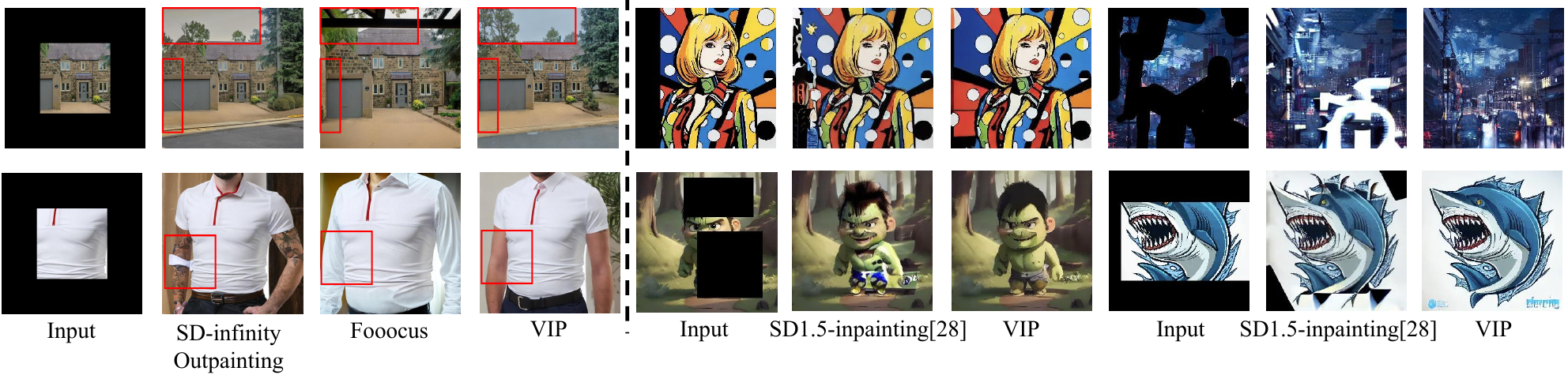}
  \caption{Left: Visualization comparison between more open-source outpainting methods; Right: Qualitative comparison on more types of images.}
  \label{fig_other}
\end{figure}

To show the superiority of VIP, we also compare it to Fooocus \footnote{https://github.com/lllyasviel/Fooocus} as shown in Fig. \ref{fig_other}, which is also widely used in the community. It can be found that VIP can achieve better performance than the two methods. For example, Fooocus frequently outpaints elements that do not match the original region.

\begin{figure}[t]
  \centering
  \includegraphics[width=\textwidth]{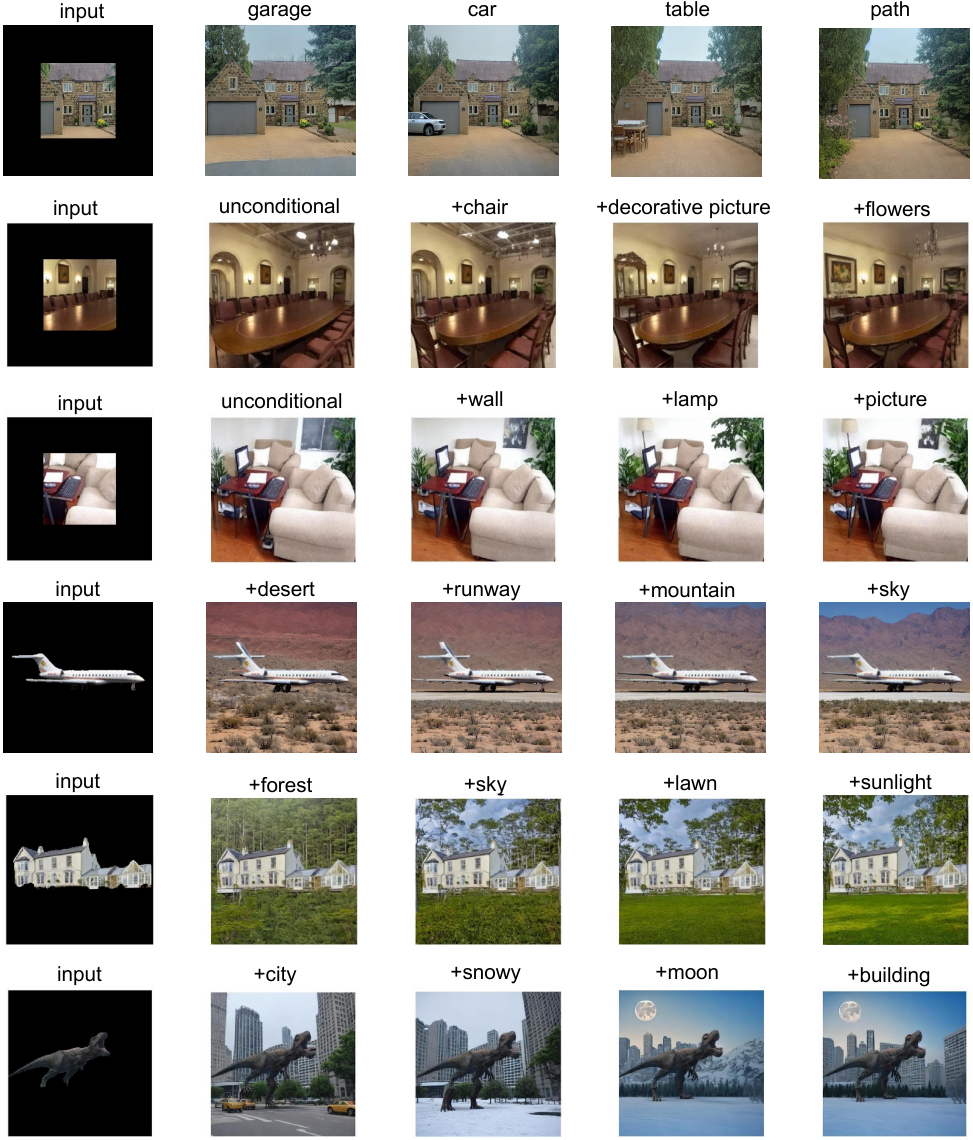}
  \caption{Visualization of more customized outpainting results.}
  \label{fig_conditional}
\end{figure}

In addition, under various outpainting scenes, more conditional customized results obtained on samples from the Laion-2B dataset are also shown in Fig. \ref{fig_conditional}. It can be seen that with various prompts, we can obtain visually plausible customized outpainting results. 

\end{document}